%% file: main.tex
\documentclass{article}

\usepackage{microtype}
\usepackage{graphicx}
\usepackage{subfigure}
\usepackage{booktabs} 

\usepackage{hyperref}
\usepackage{times}
\input{math_commands.tex}

\usepackage[utf8]{inputenc} 
\usepackage[T1]{fontenc} 
\usepackage{hyperref} 
\usepackage{url} 
\usepackage{booktabs} 
\usepackage{amsfonts} 
\usepackage{nicefrac} 

\usepackage{rotating}
\usepackage{subfigure}
\usepackage{cleveref}
\usepackage{amsmath}
\usepackage{amssymb}
\usepackage{mathtools}

\usepackage[accepted]{preprint}

\icmltitlerunning{Aggregating explanation methods}

\begin{document}

\twocolumn[
\icmltitle{Aggregating explanation methods for stable and robust explainability}

\begin{icmlauthorlist}
	\icmlauthor{Laura Rieger}{dtu}
	\icmlauthor{ Lars Kai Hansen}{dtu}
\end{icmlauthorlist}

\icmlcorrespondingauthor{Laura Rieger}{lauri@dtu.dk}

\icmlaffiliation{dtu}{DTU Compute, Technical University Denmark, 2800 Kgs. Lyngby, Denmark}

\icmlkeywords{Machine Learning, ICML}

\vskip 0.3in
]
\printAffiliationsAndNotice{} 

\begin{abstract}
	Despite a growing literature on explaining neural networks, no consensus has been reached on how to explain a neural network decision or how to evaluate an explanation.
Our contributions in this paper are twofold. First, we investigate schemes to combine explanation methods and reduce model uncertainty to obtain a single aggregated explanation. We provide evidence that the aggregation is better at identifying important features, than on individual methods.
Adversarial attacks on explanations is a recent active research topic. As our second contribution, we present evidence that aggregate explanations are much more robust to attacks than individual explanation methods. 
\end{abstract}

\section{Introduction}

Despite the great success of neural networks especially in classic visual recognition problems, explaining the networks' decisions remains an open research problem \cite{samekexplainable}. 
This is due in part to the complexity of the visual recognition problem and in part to the basic 'ill-posedness' of the explanation task.
This challenge is amplified by the fact that there is no agreement on what a sufficient explanation is and how to evaluate an explanation method.

Many different explanation strategies and methods have been proposed \citep{Simonyan2013, Zeiler2014, Bach2015,Selvaraju, Smilkov2017a, Sundararajan2017a}. Focusing on visual explanations for individual decisions, most methods either use a backpropagation approach or aim to construct a simpler linear model with an intuitive explanation.
The plethora of explanation approaches is a signature of the high-level epistemic uncertainty of the explanation task.

This paper is motivated by a key insight in machine learning: Ensemble models can reduce both bias and variance compared to applying a single model. A related approach was pursued for \textit{functional} visualization in neuroimaging \citep{hansen2001consensus}. Here we for the first time explore the potential of aggregating explanations of individual visual decisions in reducing epistemic uncertainty for neural networks.

We test the hypothesis that ensembles of multiple explanation methods are more robust than any single method.
This idea is analyzed theoretically and evaluated empirically. We discuss the properties of the aggregate explanations and provide visual evidence that they combine features, hence are more complete and less biased than individual schemes. Based on this insight, we propose two ways to aggregate explanation methods, \textit{AGG-Mean} and \textit{AGG-Var}. In experiments on Imagenet, MNIST, and FashionMNIST, the aggregates identify relevant parts of the image more accurately than any single method and are more robust to adversarial attacks.

\section{Related Work}
\label{sec:rel_work}
\subsection{Explanation methods}
The open problem of explainability is reflected in a lot of recent work \citep{Kindermans,Selvaraju,Bach2015,Zhang,Zhou2015,Ancona2017,Ribeiro,Rieger2018,Kim2017,lundberg2017unified,zintgraf2017visualizing, Simonyan2013, Zeiler2014,Selvaraju, Smilkov2017a, Sundararajan2017a,shrikumar2017learning,montavon2017explaining,chang2018explaining}. We focus on generating visual explanations for single samples. To our knowledge the first work in this direction was 
\citet{Simonyan2013} with \textit{Saliency Maps (SM)} that proposed backpropagating the output onto the input to gain an understanding of a neural network decision. The relevance for each input dimension is extracted by taking the gradient of the output w.\ r.\ t.\ to the input. This idea was extended by \citet{Springenberg2014} into \textit{Guided Backpropagation (GM)} by applying ReLU non-linearities after each layer during the backpropagation. Compared to Saliency, this removes visual noise in the explanation. 
\textit{Grad-CAM (GC)} from \citet{Selvaraju} is an explanation method, developed for use with convolutional neural networks. By backpropagating relevance through the dense layers and up-sampling the evidence for the convolutional part of the network, the method obtains coarse heatmaps that highlight relevant parts of the input image. 
\textit{Integrated Gradients (IG)} \citet{Sundararajan2017a} sums up the gradients from linearly interpolated pictures between a baseline, e.g.\ a black image, and the actual image.	
\textit{SmoothGrad (SG)} filters out noise from a basic saliency map by creating many samples of the original input with Gaussian noise \cite{Smilkov2017a}. The final saliency map is the average over all samples. 

Finally, we also consider \textit{LIME} \cite{Ribeiro}. In contrast to the other methods, \textit{LIME} is not based on backpropagation. Instead, it approximates the neural network with a linear model locally around the input to be explained. The coefficients of the linear model for the respective input dimensions give the importance of each dimension. Compared to the other methods this is much more computationally expensive as it requires many passes through the neural network. 

Considering the plethora of explanation methods, evaluating them objectively has been a topic of recent work. In this paper we will evaluate explanation methods using two approaches: 

\citet{Ancona2017} proposed a different approach to evaluate explanation methods, called {Sensitivity-\textit{n}}, based on the notion that the decrease in output when a number of inputs are canceled out should be equal to the sum of their relevances. For a range of $ n $ (between 1 and the total number of inputs) they sample a hundred subsets of the input. For each $ n $, the Pearson Correlation Coefficient (PCC) between the decrease in output, when the subset of features is removed, and the sum of their relevances is reported. The result is a curve of the PCC dependent on the percentage of the input being removed. For a good explanation method, the PCC will decrease slowly. 

We will also use a variant of the method proposed by \cite{samek2016evaluating} as introduced in \cite{anonym}. High-dimensional images get divided up into squares. Squares with high relevance (as measured by the explanation method to be evaluated) consecutively get replaced with a non-informative baseline. The difference between the original output and the output for the iteratively degraded images indicates the quality of the explanation method. For good explanation methods that accurately identify areas of high relevance, the output will go down earlier as relevant areas are replaced earlier. We calculate the area over this curve. For details we refer to \cite{anonym}, which is provided in the supplements (currently under review elsewhere).

\subsection{Adversarial attacks on explanation methods}
While adversarial examples for classification are well-known, recently there has been growing interest in adversarial manipulation of explanations \cite{ghorbani2019interpretation,heo2019fooling,dombrowski2019explanations}.
Attacks on explanation can serve multiple purposes including "fairwashing" \cite{aivodji2019fairwashing}. All of these methods exploit the fully differentiable nature of neural networks and iteratively update the input (or the model weights) to change the explanation while only minimally changing the input and output. The goal is to manipulate the explanation while keeping the input and output (visually) similar. It is assumed that the network architecture and weights are known and that either the input (\cite{ghorbani2019interpretation,zhang2018interpretable,dombrowski2019explanations}) or the network weights \cite{heo2019fooling} can be changed by the attacker.

Focussing on changing the input,
 \citep{ghorbani2019interpretation},\citep{zhang2018interpretable} and \citep{dombrowski2019explanations} attack the explanation by manipulating the image, not changing the network weights. Interestingly, \citep{zhang2018interpretable} discuss the transferability of attacks and conclude that attacks are not that transferable. If the attacker is allowed to modify networks weights, as in \cite{heo2019fooling}, the attacks generalize to all the considered explanation methods.
 
 \citep{zhang2018interpretable} also discuss various strategies where the adversarial attacks the label without modifying the explanation and vice versa. The former is interesting if the explanations is used to validate the decision.
 
 Here we are interested in the situation where the attacker can modify the input but not the network. We investigate the transferability, c.f., \citep{zhang2018interpretable}, and hypothesise that the limited transferability leads to improved robustness of the ensemble explanation.
 While ensemble methods have been proposed earlier as a defense for attacks on the label \cite{tramer2017ensemble}, they have not previously been investigated as a defense mechanism against attacks on explanations.
 
\section{Aggregating explanation methods to reduce variance}

\label{sec:agg}
All currently available explanation methods have weaknesses that are inherent to the approach and include significant uncertainty in the resulting heatmap \citep{Kindermans, Adebayo, Smilkov2017a}.
A natural way to mitigate this issue and reduce noise is to combine multiple explanation methods. Ensemble methods have been used for a long time to reduce the variance and bias of machine learning models. We apply the same idea to explanation methods and build an ensemble of explanation methods.

We assume a neural network $ F: X \mapsto y $ with $ X \in \R^{m \times m} $ and a set of explanation methods $ \{e_j\}_{j=1}^{J} $ with $ e_j : X,y,F \mapsto E$ with $ E \in \R^{m \times m} $. We write $ E_{j,n} $ for the explanation obtained for $ X_n $ with method $ e_j $ and denote the mean aggregate explanation as $ \bar{e} $ with $\bar{E}_n = \frac{1}{J}\sum_{j=1}^J E_{j,n}$. While we assume the input to be an image $ \in \R^{m \times m} $, this method is generalizable to inputs of other dimensionalities as well.

To get a theoretical understanding of the benefit of aggregation, we hypothesize the existence of a 'true' explanation $ \hat{E} _n$. This allows us to quantify the error of an explanation method as the mean squared difference between the 'true' explanation and an explanation procured by an explanation method, i.e.\ the MSE. 

For clarity we subsequently omit the notation for the neural network. We write the error of explanation method $ j $ on image $X_n$ as
$ \text{err}_{j,n}= || {E}_{j,n} - \hat{E}_n||^2 $
with \[ \text{MSE}(E_j) = \frac{1}{N} \sum_{n} \text{err}_{j,n} \]
and $ \text{MSE}(\bar{E})= \frac{1}{N}\sum_{n} || {\bar{E}}_{n} - \hat{E}_n||^2$ is the MSE of the aggregate. 
The typical error of an explanation method is the mean error over all explanation methods \[ \overline{\text{MSE}} =\frac{1}{J}\sum_j \text{MSE}(E_j). \]
With these definitions we can do a standard bias-variance decomposition \citep{geman1992neural}. Accordingly we can show the error of the aggregate will be less that the typical error of explanation methods,
\begin{eqnarray} 
\overline{\text{MSE}} = &\frac{1}{N}\sum_{n} \frac{1}{J}\sum_{j}||\hat{E}_n - E_{j,n} ||^2 \\
= &\frac{1}{N}\sum_{n} ||\hat{E}_n - \bar{E}_n ||^2 \\
&+ \frac{1}{NJ}\sum_{n,j}|| \bar{E}_n - E_{j,n} ||^2 ,\nonumber
\end{eqnarray}
hence,
\begin{eqnarray}
\label{eq:specific_methods}
\overline{\text{MSE}} = &\frac{1}{J}\sum_{j} \frac{1}{N}\sum_{n} \underbrace{||\bar{E}_n - E_{j,n}||^2}_{\text{epistemic uncertainty}} + \text{MSE}(\bar{E})\\
 \geq&\hspace{-45mm}\text{MSE}(\bar{E}). \nonumber
\end{eqnarray}
A detailed calculation is given in the appendix.
The error of the aggregate $\text{MSE}(\bar{E}) $ is less than the typical error of the participating methods. The difference - a `variance' term - represents the epistemic uncertainty and only vanishes if all methods produce identical maps.
By taking the average over all available explanation methods, we reduce the variance of the explanation compared to using a single method. To obtain this average, we normalize all input heatmaps such that the relevance over all pixels sum up to one. This reflects our initial assumption that all individual explanation methods are equally good estimators. We refer to this approach as \textit{{AGG-Mean}}.
\begin{equation} \label{eq:aggmean}
E_{\text{Agg-Mean},n} =\frac{1}{J} \sum_{j=1}^{J} E_{j,n} 
 \end{equation}

This estimator however does not take into account the estimate of the local epistemic uncertainty, i.e.\ the disagreement between methods. A way to incorporate this information is to form an 'effect size' map by dividing 
the mean aggregate locally with its standard deviation \citep{sigurdsson2004detection}. Intuitively, this will assign less relevance to segments with high disagreement between methods.

For stability, we divide not directly by the local variance but add a constant $ \epsilon $ to the estimate of the local variance. This can be interpreted as a smoothing regularizer or a priori information regarding epistemic and aleatoric uncertainties. We refer to this approach as \textit{AGG-Var}.
\begin{equation}\label{eq:aggvar}
 E_{\text{AGG-Var}, n} = \frac{1}{J} \sum_{j=1}^{J} \frac{E_{j,n} }{\sigma(E_{j \in J,n }) + \epsilon}
\end{equation}
where $ \sigma(E_{j \in J,n }) $ is the point-wise standard deviation over all explanations $ j \in J $ for $ X_n $

\section{Aggregating explanation methods to reduce vulnerability}
With the increasing interest and practical importance of explainability of neural networks the interest in methods for manipulation and control of explanations is also increasing. A typical scenario is to make impercetible changes to the input of the neural network such that the output/label is unchanged while the explanation changes according to a given goal. Such effort could, e.g., be used to hide bias or other fairness issues a given classifier might have . 

\citep{dombrowski2019explanations} showed that explanations can be made more robust by replacing the ReLU nonlinearity with a Softplus function. However, this requires changing the network and using a different architecture for classification and explanation, which is highly undesireable as it defeats the purpose of the explanation. The analysis of \citep{zhang2018interpretable} showed that transferability of attacks is limited, hence, our ensemble of multiple explanations may offer robustness also towards certain types of adversarials. 

In the following we will assume an attacker who has full knowledge of the neural network, including the architecture and weights. In contrast to \citep{heo2019fooling}, however, we will assume that the attacker cannot \emph{change} the neural network, following \cite{dombrowski2019explanations,ghorbani2019interpretation}. Furthermore the attacker has full control over the input to the neural network. The goal is to adversarially manipulate the image to according to a predefined objective. 

In the following, $ x $ will refer to the original image. $ \hat{x} $ is the 'target' image. The objective is to produce an adversarial input $ x' $ with $ x' \approx x $ but the explanation $ E_{x'} \approx E_{\hat{x}}$. While we focus on assimilating the explanation map of another input as in \citep{dombrowski2019explanations}, all techniques introduced can be readily adapted to other objectives, f.e.\ to move the mass center of the explanation. 

Exploring the robustness of aggregates of multiple explanation methods we concentrate on the following two scenarios:
\paragraph{Arsenal of explanation methods}
In this scenario we have a pool of potential explanation methods. The attacker does not have knowledge of what explanation method is used and optimizes for a different explanation method than is used by the defender. 
The success of the attack depends on how readily an attack of one explanation method translates to another method. 

We hypothesize that attacks on explanation methods are fragile and do not translate well across explanation methods, as they exploit locally high variances in the gradient landscape. This hypothesis is examined in \cref{ssub:attack}.
\paragraph{Aggregation of explanation methods}
In this more challenging scenario we aggregate multiple explanation methods as described in \cref{eq:aggmean}. The attacker knows the exact explanation methods and ratio going into the mixture and attacks this aggregation.

Many attribution-based methods are utilizing the gradient $ \frac{\delta y}{\delta x}(x) $ of the output to create an explanation. Due to the non-linearity of the neural network, the gradient can change rapidly with small distances in input space \cite{dombrowski2019explanations,ghorbani2019interpretation}. Attacks on explanation methods exploit this vulnerability.

AGG-Var as defined in \cref{eq:aggvar} will not be considered as a defence against adversarial manipulation since training with an objective containing a variable in both the nominator and denominator is computationally undesirable. While this is an issue for the attack, it also makes evaluation considerably harder.
\section{Experiments}
\label{sec:experiments}
We first evaluate aggregations of explanation techniques against vanilla techniques with a variation of the evaluation method proposed in \cite{Bach2015}, Sensitivity-\textit{n} and qualitative evaluation. Subsequently we evaluate the effectiveness of aggregating methods against adversarial attacks on explanation methods in \cref{ssub:attack}. 
In the appendix we compare aggregated methods on a dataset of human-annotated heatmaps.

\subsection{Experimental details}
We tested our method on five neural network architectures that were pre-trained on ImageNet: VGG19, Xception, Inception, ResNet50 and ResNet101 \citep{DengJ.andDongW.andSocherR.andLiL.-J.andLiK.andFei-Fei2009,simonyan2014very,he2016deep,chollet2017xception,szegedy2016rethinking}.
\footnote{Models retrieved from \href{https://github.com/keras-team/keras}{{https://github.com/keras-team/keras}}.}
Additionally, we ran experiments on CNN trained on the MNIST and FashionMNIST dataset \cite{mnist, Xiao2017}.
\subsection{Quantitative evaluation}
To quantitatively compare the quality of the explanation methods on a more challenging dataset, we use an evaluation method similar to the approach described in \citep{samek2016evaluating} (see \cref{sec:rel_work}). Instead of dividing the image into squares, we use an established segmentation algorithm, SLIC, to divide the image into semantically meaningful areas \cite{achanta2012slic}. This makes the information that is being removed each step more meaningful and reduces variance in the result \cite{anonym}
 A manuscript with a detailed description of this approach, including an evaluation of evaluation methods is currently under review elsewhere and included in the supplements \cite{anonym}. 

We compared the aggregation methods against Saliency (SM), Guided Backpropagation (GB), SmoothGrad (SG), Grad-CAM (GC) and Integrated Gradients (IG) to have a selection of attribution-based methods. Additionally we compared against LIME as a method that is not based on attribution but rather on local approximation \cite{Ribeiro}. The aggregations are based on all attribution-based methods.

Some of the methods result in positive and negative evidence. We only considered positive evidence for the ImageNet tasks to compare methods against each other. To check that this does not corrupt the methods, we compared the methods that do contain negative results against their filtered version and found negligible difference between the two versions of a method in the used metrics.
For \textit{AGG-Var} we introduced an additional stabilizing parameter, $ \epsilon $ to the divisor. In our experiments we set $ \epsilon $ to be ten times the mean $ \sigma $ over the entire dataset.

\label{subsec:segment}

\begin{table*}[!hbt]
	\caption{Scores across methods and architectures. \textit{AGG-Mean} and \textit{AGG-Var} surpass all methods in all scenarios. All SE $< 0.05 $. }
	\label{tab:results}
	
	\begin{center}
		\begin{small}
			\begin{sc}
				\input{tables/AUC}
			\end{sc}
		\end{small}
	\end{center}
\end{table*}

In \cref{tab:results} we show the results of this evaluation. High scores indicate that the explanation method correctly labels parts of the image that are relevant for the classification as important.
We include two non-informative baselines. \textit{Random} randomly chooses segments to remove. 
\textit{Sobel} is a sobel edge detector. Neither of them contain information about the neural network.

All explanation methods have a higher evaluation score than the random baseline on all architectures tested, indicating that all methods contain information about the image classification. Except for LIME, all methods also surpass the stronger baseline, \textit{SOBEL}. 
The ranking of unaggregated methods varies considerably between architectures. This variance indicates that the accuracy of an explanation method depends on the complexity and structure of the neural network architecture.
For all architectures \textit{AGG-Mean} and \textit{AGG-Var} have a higher score than any non-aggregated method. For ResNet101 the difference between the best unaggregated method and the aggregated methods is especially large. We hypothesize, that the benefit of aggregating explanation methods increases for more complex neural network with large epistemic uncertainty on the explanation. 
We empirically confirmed that aggregating methods improves over unaggregated methods and more reliably identifies parts of the images that are relevant for classification.

\subsection{Qualitative visual evaluation}
\label{subsec:qual_eval}
\begin{figure*}[!bht]
	\begin{center}
		\includegraphics[width=.9\linewidth]{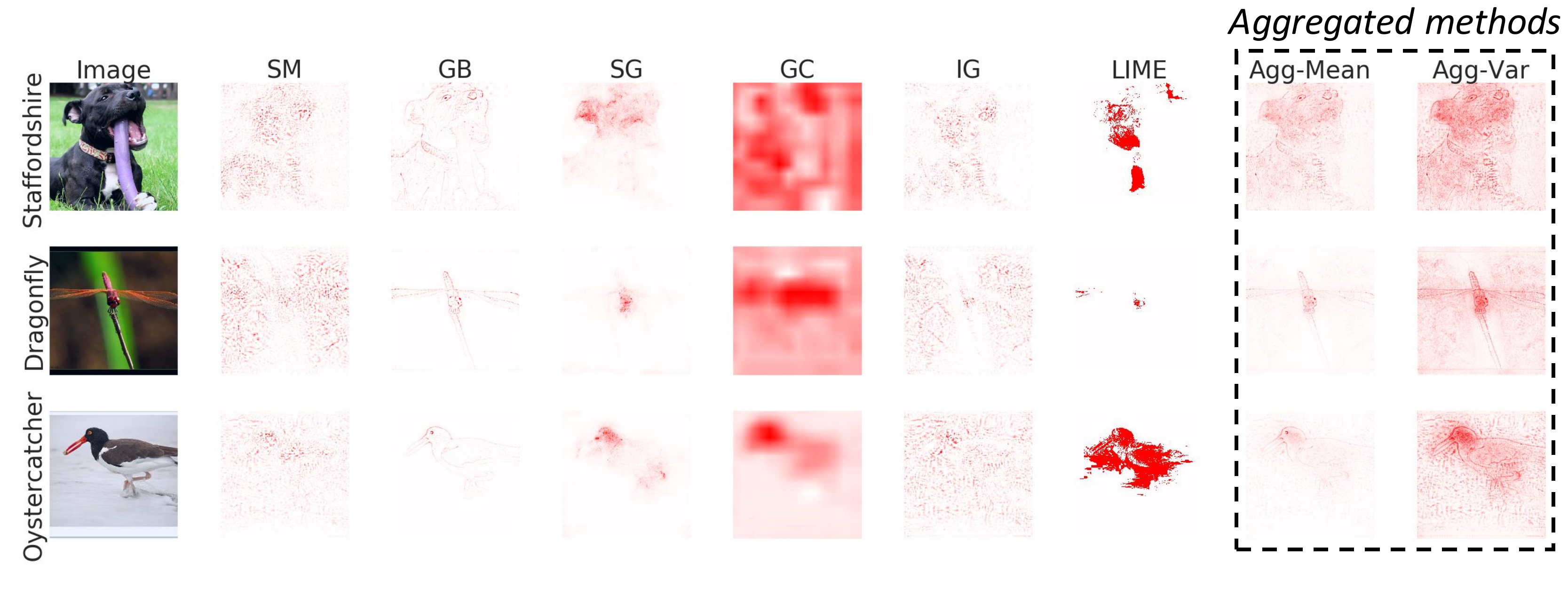}
	\end{center}
	\caption{Example images from Imagenet and the heatmaps produced by different methods on VGG19. Aggregated methods combine features from all methods. Too heavy focus on one feature by SmoothGrad is smoothed away by the aggregation. }
	\label{fig:example_img}
\end{figure*}

We show heatmaps for individual examination for each of the methods in \cref{fig:example_img} and compare qualitatively.
While visual evaluation of explanations for neural networks can be misleading, there is no better way available of checking whether any given explanation method agrees with intuitive human understanding \citep{Adebayo}. Additionally, we compute alignment between human-annotated images and the explanation methods in the appendix, using the human benchmark for evaluation introduced in \cite{Mohseni2018}.

\textit{AGG-Var} combines features of the aggregrated methods by attributing relevance on the classified object as a whole, but considering smaller details such as the face of an animal as more relevant. It is a combination of the detail-oriented and shape-oriented methods. 
Compared to SmoothGrad, which concentrates on one isolated feature, the relevance is more evenly distributed and aligned with our human intuition that classification takes context into account and does not rely on e.g.\ only the beak of a bird.
We conclude that combining explainability methods provides a meaningful visual improvement over single methods.

\subsection{Evaluation with Sensitivity-\textit{n} on low-dimensional input}
\label{subsec:evalsens}
To quantitatively compare explanation methods on a low-dimensional input we use Sensitivity-\textit{n} \citep{Ancona2017}. The exact procedure is described in \cref{sec:rel_work}. We compare on MNIST \citep{mnist} and FashionMNIST \citep{Xiao2017}, two low-dimensional dataset with a basic CNN\footnote{Model and code retrieved from \href{https://github.com/keras-team/keras/blob/master/examples/mnist\_cnn.py}{https://github.com/keras-team/keras/blob/master/examples/mnist\_cnn.py}.} (architecture in appendix) . 
We follow the procedure suggested in \cite{Ancona2017} and test on a hundred randomly sampled subsets for 1000 randomly sampled test images. The number of pixels in the set \textit{n} is chosen at fifteen points logarithmically spaced between $ 10 $ and $ 780 $ pixels. 

As described in \cref{sec:rel_work}, for a range of $ n $ (between 1 and the total number of inputs) a hundred subsets of the input features are removed. For each $ n $, the average Pearson Correlation Coefficient (PCC) between the decrease in output and the relevance of the removed output features is reported. The result is a curve of the PCC dependent on the removed percentage. 

\begin{figure*}[!bt]
	\begin{center}
		\includegraphics[width=.8\linewidth]{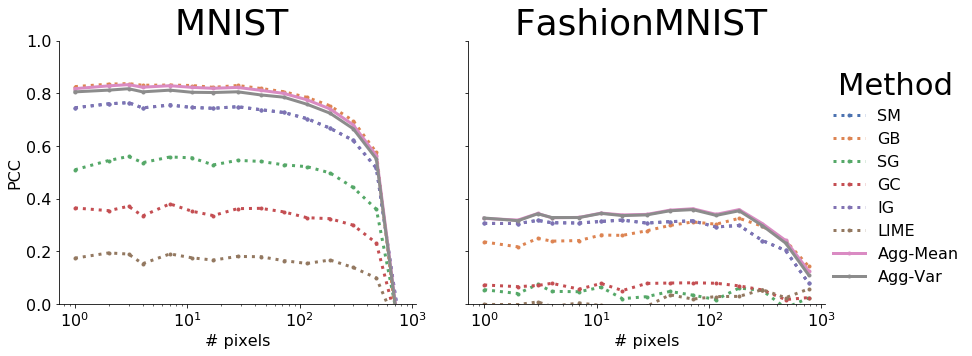}
	\end{center}
	\caption{Sensitivity-\textit{n} for explanation methods. Higher is better. The proposed methods, \textit{AGG-Mean} and \textit{AGG-Var} perform better or equally good as all other methods.}
	\label{fig:nsensitivity}
\end{figure*}

We show results in \cref{fig:nsensitivity}. \textit{AGG-Mean} and \textit{AGG-Var} perform in range of the best methods. For the FashionMNIST CNN \textit{AGG-Mean} and \textit{AGG-Var} perform better than unaggregated methods. For the MNIST CNN Guided Backpropagation and \textit{AGG-Mean} perform best. Considering that MNIST has a nearly binary data representation (black/white), 'removing' a pixel by setting it to black is informative in itself, making Sensitivity-\textit{n} less reliable for networks trained on this particular dataset.
For both networks (trained on FashionMNIST and MNIST respectively), SmoothGrad and GradCAM perform considerably worse than the other methods. 

In summary, aggregation seems to not be as beneficial when applied to a low-dimensional, "easier" tasks such as MNIST as it is for ImageNet, performing in range of the best unaggregated method. We hypothesize that this is because there is less epistemic uncertainty in explanations for less complex tasks and network architectures.

\subsection{Attacking explanation methods}
\label{ssub:attack}
We evaluate how robust aggregated methods are against adversarial attacks, compared to unaggregated methods. In all cases we assume that the attacker has full knowledge of the network architecture and weights (white box attack) but cannot change them. However, the attacker has full control over the input. 

Following \cite{dombrowski2019explanations} we run experiments on a pretrained VGG16 \cite{simonyan2014very} and consider Layerwise Relevance Propagation (LRP), Saliency Mapping (SM), Guided Backprop and Integrated Gradients. The latter was not used in the aggregation. Unless otherwise noted we followed \citep{dombrowski2019explanations} in the choice of hyperparameters for attacking explanation methods. Since the ReLU function used in neural networks is not twice differentiable, we replace it with a differentiable approximation, SoftPlus for the iterative creation of the adversarial input. The final manipulated heatmaps are created with the ReLU non-linearity. Further details about the experiments are in the appendix.

We consider the two scenarios introduced in \cref{ssub:attack}. In all cases, the objective of the attacker is to make the explanation of input $ E_{x'} \approx E_{\hat{x}} $ while keeping $ x' \approx x $. To do this, he manipulates $ x' $.

We visually confirmed that the adversarial images look similar to the input images and provide examples in the appendix. We measure the difference between the start explanation $ E_x $ target explanation $ E_{\hat{x}} $ and adversarial explanation $ E_{x'} $ with the MSE (Mean Square Error), the PCC (Pearson Correlation Coefficient) and the top-\textit{k} intersection with k being ad-hoc set to 10\% \citep{dombrowski2019explanations,Ghorbani2017}. 

 In all metrics, explanations obtained with different methods have different 'base' values (similarity between the explanations of two randomly chosen images) due to structural differences between explanation methods. To account for this, we consider for each similarity metric $ m_{\text{sim}} $ the difference $ m_{\text{sim}}(E_{\hat{x}}, E_{x'}) - m_{\text{sim}}(E_{\hat{x}}, E_x)$, i.e.\ how much \textit{more} similar the attack makes $ E_{x'} $ look to $ E_{\hat{x}} $. For the MSE, this results in a negative score, since the difference between the target and the attack is less than between the target and the starting point. For all metrics, the ideal score is $ 0 $, i.e.\ the attack did not change the explanation at all. Thus, for MSE a high value is desirable, for PCC and top-\textit{k} union a low value is desirable.

\paragraph{Transferability of attacks on explanation methods}

We consider a case where the attacker does not know what explanation method is used, i.e.\ we attack a different explanation method than the one that is used. If the attack translates well, i.e.\ the image manipulation fools both methods, similarity metrics should be similar for both explanation methods. 

\begin{figure}[!ht]
	\begin{center}
		\includegraphics[width=\linewidth]{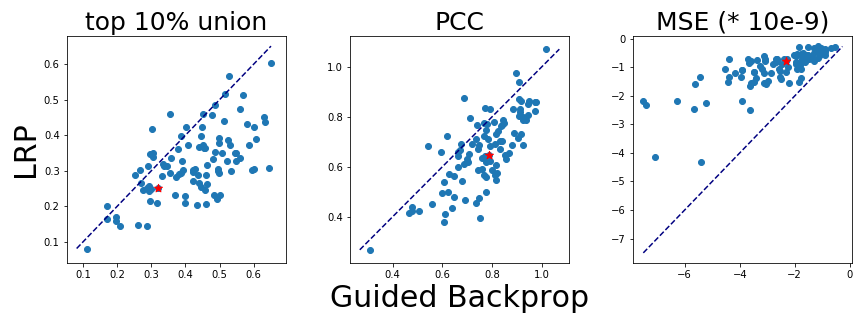}
	\end{center}
	\caption{ Attacking one method does not translate to attacks on the other methods. Similarity metrics (topK and PCC) should be low, MSE should be high. The red dot is the sample visualized in \cref{fig:attack_example_pumpkin} }
	\label{fig:adv_attack}
\end{figure}

In \cref{fig:adv_attack} we provide results for attacking Guided Backpropagation and extracting an explanation with LRP. For a hundred samples we visualize for each sample the respective similarity metrics for both explanation methods in \cref{fig:adv_attack}. If the attack translates well, the points should lie on the identity line in \cref{fig:adv_attack}. Samples below the identity line for PCC and topK and above for MSE indicate that the attack does not generalize to other explanation methods.

\begin{figure}[!hbt]
	\begin{center}
	\includegraphics[width=\linewidth]{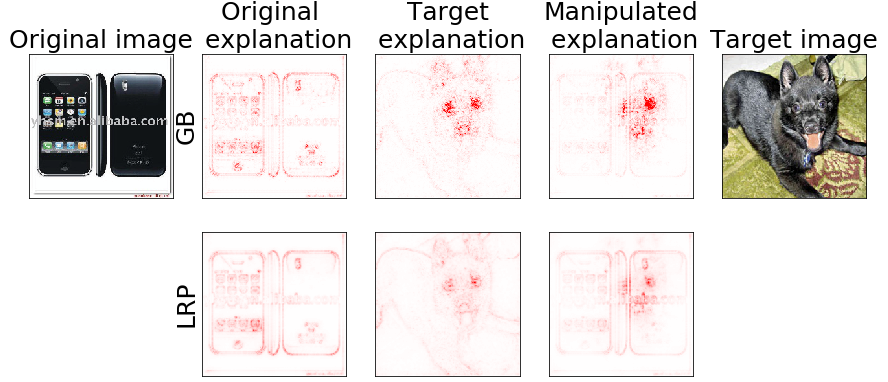}
	\end{center}
	\caption{Example showing the transferability of attacks. The adversarial input was calculated to attack GB (upper row). We then extracted the explanation with LRP (lower row). The attack does not transfer well to LRP. 
		To visualize details better we clipped values at the 99th percentile. }
	\label{fig:attack_example_pumpkin}
\end{figure}

As visible in \cref{fig:adv_attack} and anecdotally in \cref{fig:attack_example_pumpkin} (red data point in \cref{fig:adv_attack}), attacks perform much worse on other methods (here LRP) than the targeted one (here GB). We provide statistics for other combinations in the appendix. 

The lack of transferability results are in line with the findings of \citep{zhang2018interpretable}. 

\paragraph{Attacking aggregations of explanation methods.}
\label{subsec:attack_eval}

In the second scenario the attacker knows that the explanations are aggregated and attacks the aggregation. We aggregate LRP, GB and SM and compare against those methods as well as Integrated Gradient. IG was not included in the aggregation as it requires sampling for each step, making it computationally much more expensive than the other methods. 

\begin{figure}[!hbt]
	\begin{center}
		\includegraphics[width=\linewidth]{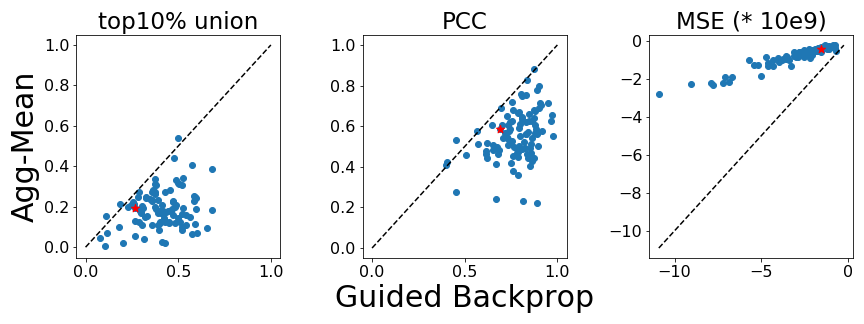}
	\end{center}
	\caption{Visualizing the success of an attack on AGG-Mean compared to unaggregated method (Guided Backprop). Similarity metrics (topK and PCC) should be low, MSE should be high for less similarity between target and adversarial. In the majority of cases, AGG-Mean is more robust than Guided Backprop. The red dot is the sample visualized in \cref{fig:adv_attack_agg} }
	\label{fig:adv_attack_agg_summary}
\end{figure}

\begin{figure*}[hbt]
	\begin{center}
		\includegraphics[width=.7\linewidth]{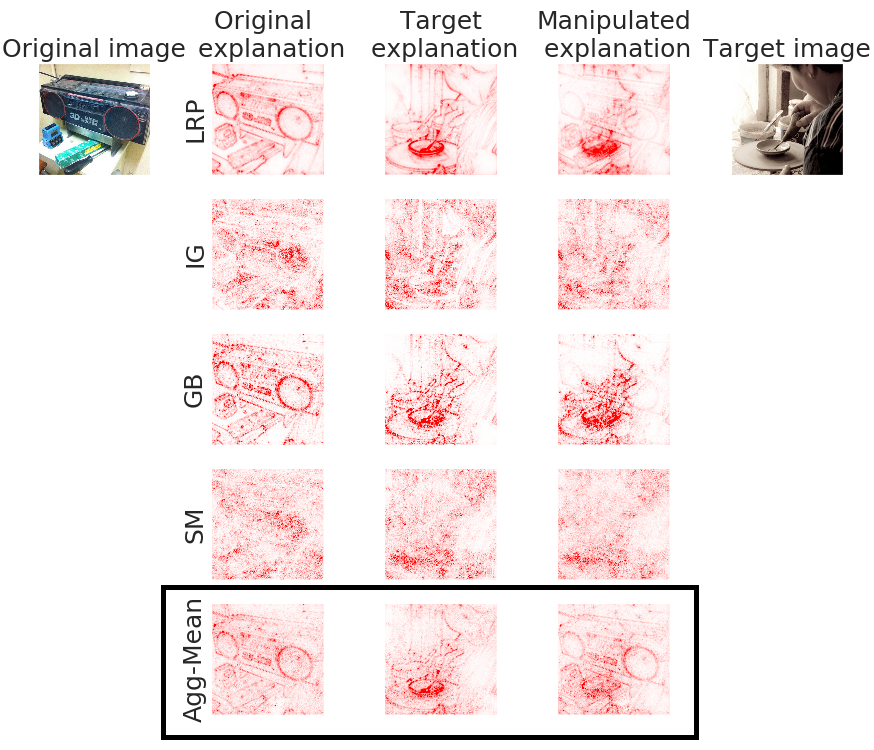}
	\end{center}
	\caption{Attacking explanation methods.  We provide the adversarial input images in the appendix. AGG-Mean visibly preserves original explanation best, thus being the most resistant to adversarial attacks. To visualize details better we clipped values at the 99 percentile.}
	\label{fig:adv_attack_agg}
\end{figure*}

In \cref{tab:results_attacks} we provide  metrics averaged over a hundred samples.  Agg-Mean outperforms unaggregated methods. We also provide a direct comparison to GuidedBackprop in \cref{fig:adv_attack_agg_summary}. To give an intuition on what differences int the  metrics  look like, we visualize a sample (red dot in \cref{fig:adv_attack_agg_summary})  in \cref{fig:adv_attack_agg}. We see that Agg-Mean opposed to the unaggregated methods largely preserves the original heatmap after the attack. More examples are provided in the appendix.

The resilience of the aggregate to attacks can be understood in terms of averaging induced smoothness. In \citep{dombrowski2019explanations}
the beneficial effects of averaging in the SmoothGrad method are described. As noted in \citep{dombrowski2019explanations} SmoothGrad is computationally expensive.
We conjecture that the diversity of the methods involved in the present aggregate implies that smoothing can be achieved at less computational effort.

\begin{table*}[!hbt]
	\caption{Evaluation scores across methods and architectures on a hundred samples. \textit{AGG-Mean} surpasses all considered methods in all metrics. Deviation is SE. }
	\label{tab:results_attacks}
	\vskip 0.15in
	\begin{center}
		\begin{small}
			\begin{sc}
				\begin{tabular}{lrrr}
					\toprule
					{} & MSE (*10e-9) & PCC & top 10\% Union \\
					method & & & \\
					\midrule
					SM 		& -0.92 	$\pm$ 0.00				& 0.74 $\pm$ 0.01 			& 0.40 $\pm$ 0.01 \\
					GB 				& -3.25	 $\pm$ 0.02			& 0.77 $\pm$ 0.01			& 0.42 $\pm$ 0.01\\
					LRP & -1.45 $\pm$ 0.01						& 0.81 $\pm$ 0.01 			& 0.49 $\pm$ 0.01 \\
					IG 				& -1.76 $\pm$ 0.01			& 0.82 $\pm$ 0.01			& 0.47 $\pm$ 0.01\\
					AGG-Mean 	& \textbf{-0.89} $\pm$ 0.01 	& \textbf{ 0.54 $\pm$ 0.01 }	& \textbf{ 0.24 $\pm$ 0.01} \\
					\bottomrule
				\end{tabular}
				
			\end{sc}
		\end{small}
	\end{center}
	\vskip -0.15in
\end{table*}

\section{Conclusion}
The explanation problem for object detection using neural networks is ill-posed: Many schemes have been proposed leaving considerable epistemic uncertainty. We propose to mitigate this by aggregating multiple explanation methods. This serves multiple purposes, reducing variance representing the epistemic uncertainty, reducing bias by extending the range of features that an explanation can highlight and reducing vulnerability to adversarial attacks. 

We gave a simple proof that aggregating explanation methods will perform at least as good as the typical individual method. In practice, we found evidence that aggregating methods outperforms any single method. This evidence was substantiated across quantitative metrics. While our results show that different vanilla explanation methods perform best on different network architectures, an aggregation supersedes all of them on any given architecture. 

Attacks on explanation has received considerable recent attention. 
We provided experimental evidence that aggregations are a more robust to adversarial manipulation than individual explanation methods. In \cite{dombrowski2019explanations} arguments are presented that the observed vulnerability is due to non-smoothness of contemporary networks. It is also argued that averaging as in SmoothGrad increases robustness. We conjectured that the averaging of the diverse set of explanation methods involved in the aggregate creates similar smoothness. We noted that in contrast to \cite{dombrowski2019explanations}, the aggregate does not require modification (smoothing) of the network.

\clearpage
\bibliography{main}
\bibliographystyle{icml2020}

\appendix
\onecolumn
\section{Appendix}
\subsection{Aggregating explanation methods to reduce variance - detailed derivation}
\label{sec:agg_appendix}
All currently available explanation methods have weaknesses that are inherent to the approach and include significant noise in the heatmap \citep{Kindermans, Adebayo, Smilkov2017a}.
A natural way to mitigate this issue and reduce noise is to combine multiple explanation methods. Ensemble methods have been used for a long time to reduce the variance and bias of machine learning models. We apply the same idea to explanation methods and build an ensemble of explanation methods.

We assume a neural network $ F: X \mapsto y $ with $ X \in \R^{m x m} $ and a set of explanation methods $ \{e_j\}_{j=1}^{J} $ with $ e_j : X,y,F \mapsto E$ with $ E \in \R^{mxm} $. We write $ E_{j,n} $ for the explanation obtained for $ X_n $ with method $ e_j $ and denote the mean aggregate explanation as $ \bar{e} $ with $\bar{E}_n = \frac{1}{J}\sum_{j=1}^J E_{j,n}$. While we assume the input to be an image $ \in R^{mxm} $, this method is generalizable to inputs of other dimensions as well.

We define the error of an explanation method as the mean squared difference between a hypothetical 'true' explanation and an explanation procured by the explanation method, i.e.\ the MSE. For this definition we assume the existence of the hypothetical 'true' explanation $ \hat{E} _n$ for image $ X_n $.

For clarity we subsequently omit the notation for the neural network.

We write the error of explanation method $ j $ on image $X_n$ as
$ err_{j,n}= || {E}_{j,n} - \hat{E}_n||^2 $
with \[ \text{MSE}(E_j) = \frac{1}{N} \sum_{n} err_{j,n} \]
and $ \text{MSE}(\bar{E})= \frac{1}{N}\sum_{n} || {\bar{E}}_{n} - \hat{E}_n||^2$ is the MSE of the aggregate. 
The typical error of an explanation method is represented by the mean

\begin{equation} 
\begin{split}
\overline{\text{MSE}} &= \frac{1}{N}\sum_{n} \frac{1}{J}\sum_{j}||\hat{E}_n - E_{j,n} ||^2 \\
&= \frac{1}{NJ}\sum_{n,j}||\hat{E}_n - E_{j,n} + \bar{E}_n - \bar{E}_n||^2 \\
&= \frac{1}{NJ}\sum_{n,j}||(\hat{E}_n - \bar{E}_n) +( \bar{E}_n - E_{j,n} ) ||^2 \\
&= \frac{1}{NJ}\sum_{n,j}|| \hat{E}_n - \bar{E}_n ||^2 + || \bar{E}_n - E_{j,n} ||^2 +\frac{1}{NJ}\sum_{n,j}\left( 2{\rm Tr}\left[(\hat{E}_n - \bar{E}_n)(\bar{E}_n - E_{j,n} )\right]\right) \\
&= \frac{1}{N}\sum_{n} ||\hat{E}_n - \bar{E}_n||^2 + \frac{1}{NJ}\sum_{n,j}|| \bar{E}_n - E_{j,n} ||^2 + 2\frac{1}{N}\sum_{n} {\rm Tr}\left[(\hat{E}_n - \bar{E}_n)\left( \frac{1}{J}\sum_{j} (\bar{E}_n - E_{j,n} )\right)\right] \\
&= \frac{1}{N}\sum_{n} ||\hat{E}_n - \bar{E}_n ||^2 + \frac{1}{NJ}\sum_{n,j}|| \bar{E}_n - E_{j,n} ||^2 + 2\frac{1}{N}\sum_{n}{\rm Tr}\left[ (\hat{E}_n - \bar{E}_n)\underbrace{\frac{1}{J}\sum_{j} (\bar{E}_n - E_{j,n} ) }_{=0} \right]\\
&= \frac{1}{N}\sum_{n} ||\hat{E}_n - \bar{E}_n ||^2 + \frac{1}{NJ}\sum_{n,j}|| \bar{E}_n - E_{j,n} ||^2 ,\nonumber
\end{split}
\end{equation}

hence,
\begin{align}
\label{eq:specific_methods2}
\overline{\text{MSE}} &=\text{MSE}(\bar{E}) + \frac{1}{NJ}\sum_{n,j} \underbrace{||\bar{E}_n - E_{j,n}||^2}_{\text{epistemic uncertainty}} \geq \text{MSE}(\bar{E}) \nonumber
\end{align}
The error of the aggregate $\text{MSE}(\bar{E}) $ is less than the typical error of the participating methods. The difference - a `variance' term - represents the epistemic uncertainty and only vanishes if all methods produce identical maps.

\subsection{Comparing aggregate of two methods}
\begin{figure}[!ht]
	\begin{center}
		\includegraphics[width=.7\linewidth]{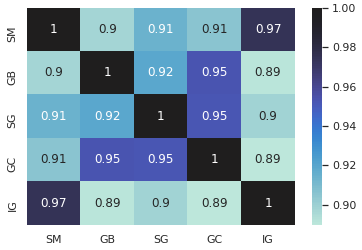}
	\end{center}
	\caption{ Ratios between the aggregate IROF and the averaged IROF of two methods. Aggregation always improves on the results, as all values outside of the diagonal are below one.}
	
	\label{fig:comparison_to_avg}
\end{figure}

In \cref{sec:agg_appendix} we showed theoretically that the average MSE of two or more explanation methods will always be higher than the error of the averaged of those methods. Empirically, we test this for the IROF score first proposed in with combinations of any two methods for ResNet101 and show the results in \cref{fig:comparison_to_avg}. For any two methods, the matrix shows the ratio between the aggregate method IROF and the average IROF of the aggregated methods. The aggregate IROF is always lower, confirming our theoretical results.

\subsection{Experimental setup }
\subsubsection{General}
We use SLIC for image segmentation due to availability and quick run time \citep{achanta2012slic}. Preliminary experiments with Quickshift showed similar results \citep{Vedaldi2008}. SLIC was chosen over Quickshift due to the quicker run time. The number of segments was set to $ 300 $ ad hoc. For a detailed description with motivation and evaluation of the evaluation metric used we refer to \cite{anonym}

For AGG-Var, we add a constant to the denominator. We set this constant to $ 10 $ times the mean std, a value chosen empirically after trying values in the range of $ [1, 10, 100] $ times the mean. \\
Evaluations were run with a set random seed for reproducibility. SE were reported either for each individual result or if they were non-significant in the caption to avoid cluttering the results. \\ 
All experiments were done on a Titan X.

\subsubsection{MNISTs}
The training for both models was equivalent. The architecture was as follows:\newline
(input)-(conv(32,3,3))-(conv(64,3,3))-(maxPool(2,2))-(dropout(0.25))\newline-(fully connected(128))-(dropout(0.5))-(output(10))

ReLU was used as a non-linearity for both. All networks were trained with Adadelta and early stopping on the validation set (patience of three epochs) \cite{zeiler2012adadelta}. The final accuracy for MNIST was 99.21\%.:
The final accuracy on FashionMNIST was 92.46\%.

\subsubsection{ImageNet}
We tested our method on five network architectures that were pre-trained on ImageNet: VGG19, Xception, Inception, ResNet50 and ResNet101 \citep{DengJ.andDongW.andSocherR.andLiL.-J.andLiK.andFei-Fei2009,simonyan2014very,he2016deep,chollet2017xception,szegedy2016rethinking}. We used the pre-trained networks VGG19, Xception and Inception, obtained from the keras library and did not change the networks in any way.
\citep{DengJ.andDongW.andSocherR.andLiL.-J.andLiK.andFei-Fei2009,szegedy2016rethinking,chollet2017xception,simonyan2014very}. 

We downloaded the data from the ImageNet Large Scale Visual Recognition Challenge website and used the validation set only. No images were excluded. The images were preprocessed to be within $ [-1,1] $ unless a custom range was used for training (indicated by the preprocess function of keras).

\subsubsection{Details about attacking explanation methods}

For a range of explanation methods we chose to compare against LRP, Gradient, Guided Backpropagation and Integrated Gradients, a range of well-known and well-established explanation methods \cite{Sundararajan2017a,Bach2015,Springenberg2014,Simonyan2013}. Since Integrated Gradients is thirty times more computationally expensive than other methods, we did not include it in the aggregation as it would have slowed down experiments considerably. 

Unless otherwise noted, all metrics are computed as the average of a hundred data samples with mean and SE. Informally, we also found that the MSE does not align well with perceived changes in the explanations, likely due to it being susceptible to outliers. 

We used a pretrained VGG16 for all experiments attacking explanation methods \cite{simonyan2014very}. 

\clearpage

\subsection{Alignment between human attribution and explanation methods}
\label{sec:human_eval}
We want to quantify whether an explanation method agrees with human judgement on which parts of an image should be important. While human annotation is expensive, there exists a benchmark for human evaluation introduced in \cite{Mohseni2018}. 
\begin{figure*}[ht]
	\begin{center}
		\includegraphics[width=\linewidth]{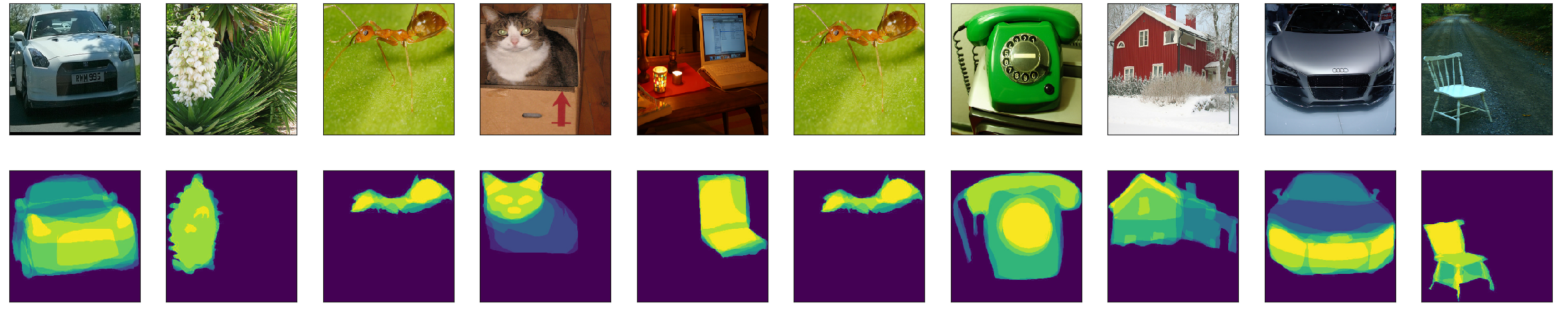}
	\end{center}
	\caption{Example images from \cite{Mohseni2018} with human-annotated overlays.}
	\label{fig:sample_human}
\end{figure*}
The benchmark includes ninety images of categories in the ImageNet Challenge (ten images were excluded due to the category not being in the ImageNet challenge) and provides annotations of relevant segments that ten human test objects found important. Example images are shown in \cref{fig:sample_human}.

While human evaluation is not a precise measure, we still expect some correlation between neural network and human judgement. 

To test the alignment, we calculate the cosine similarity,
\[\text{similarity}(e_j) = \frac{\sum_{n=1}^{N} A_n E_{j,n}}{\sqrt{\sum_{n=1}^{N}A_n^2}\sqrt{\sum_{n=1}^{N}E_{j,n}^2}}\]

between the human annotation and the explanations produced by the respective explanation methods. $ A_n $ is the human annotation of what is important for image $ X_n $

Since the images in this dataset are 224x224 pixel large, we only compute the cosine similarity for the network architectures where pretrained networks with this input size were available. 

We see that \textit{AGG-Mean} and \textit{AGG-Var} perform on-par with the best methods (SmoothGrad and GradCAM). While the aggregated methods perform better than the average explanation method, they do not surpass the best method. 

When we combine the two best-performing single methods, SmoothGrad and GradCAM, we surpass each individual method. We hypothesize that this is because the epistemic uncertainty is reduced by the aggregate.

\begin{table*}[!h]
	\caption{Cosine similarity between heatmap and human annotated benchmark. All SE below $ 0.05 $}
	\vskip 0.15in
	\begin{center}
		\begin{small}
			\begin{sc}

\input{tables/jaccard}
			\end{sc}
		\end{small}
	\end{center}
	\vskip -0.15in
\end{table*}
\clearpage

\subsection{Details about attacking explanation methods}
\paragraph{Choice of explanation methods}
We focused on explanation methods that have previously been shown to be susceptible to adversarial attacks. As such, we did not include GradCAM in the experiments, neither as a comparison or in the aggregation. 

Different explanation methods have different computational loads. Notably, SmoothGrad and IntegratedGradients involve the sampling of many explanations for a single pass, increasing computation times by the number of samples ()  and were not included in the aggregation but as a comparison. 
\paragraph{Choice of hyperparameters}
We  followed \cite{dombrowski2019explanations} for the choice of hyperparameters in learning rate and beta growth. For AGG-Mean we chose a learning rate of $ 10^{-3} $ and $ 1500 $ iterations for the attack. We heavily use their code, retrieved from the repository of the authors \footnote{https://github.com/pankessel/explanations\_can\_be\_manipulated} \cite{dombrowski2019explanations}.

\subsubsection{More examples}
\begin{figure*}[hbt]
	\begin{center}
		\includegraphics[width=\linewidth]{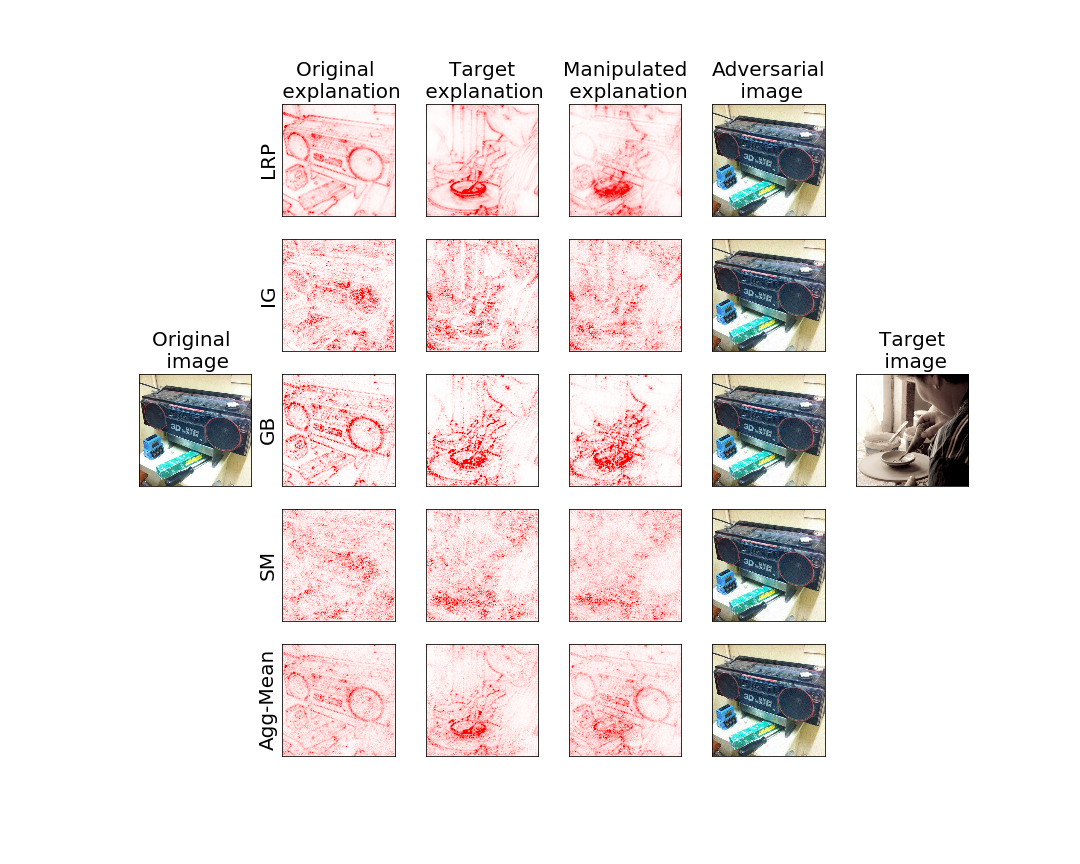}
	\end{center}
	\caption{Attack shown in the main text, including the adversarial input images. There are no visual differences for any of the adversarial inputs.}
	\label{fig:attack0}
\end{figure*}

\begin{figure*}[hbt]
	\begin{center}
		\includegraphics[width=\linewidth]{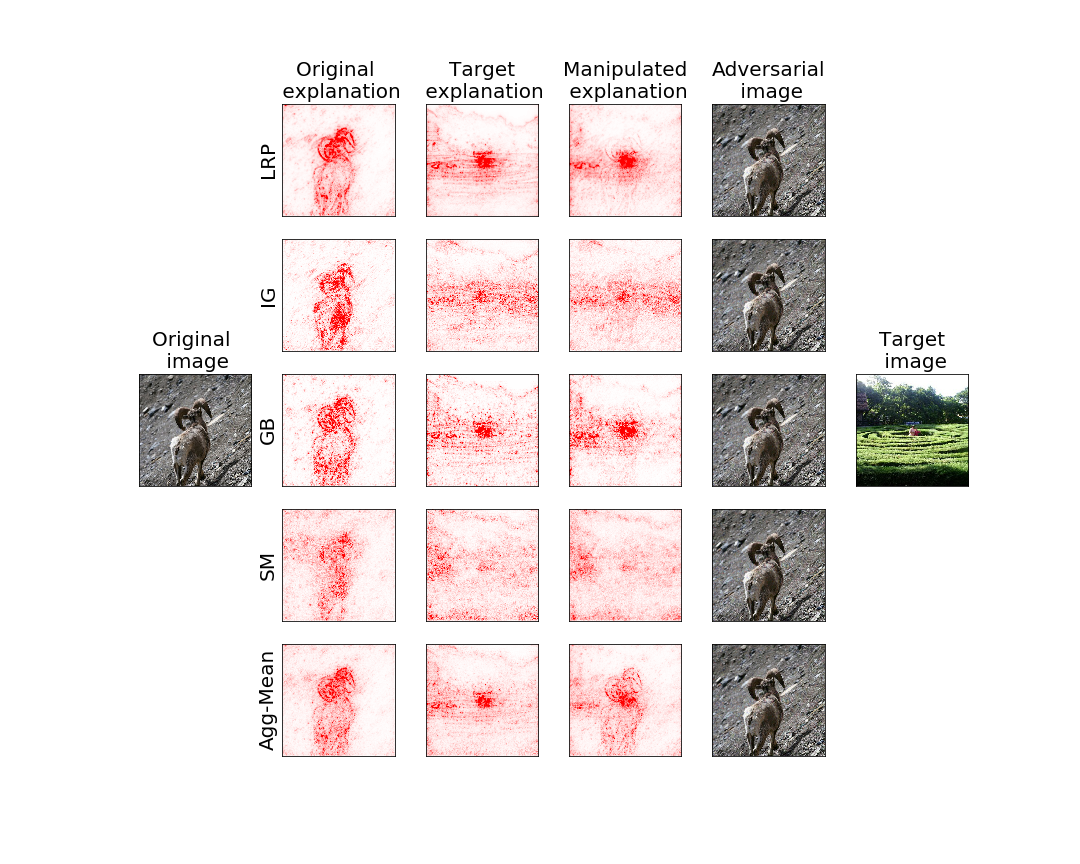}
	\end{center}
	\caption{Appendix example 1. Aggregation is more robust against attack. There are no visual differences for any of the adversarial inputs.}
	\label{fig:attack1}
\end{figure*}

\begin{figure*}[hbt]
	\begin{center}
		\includegraphics[width=\linewidth]{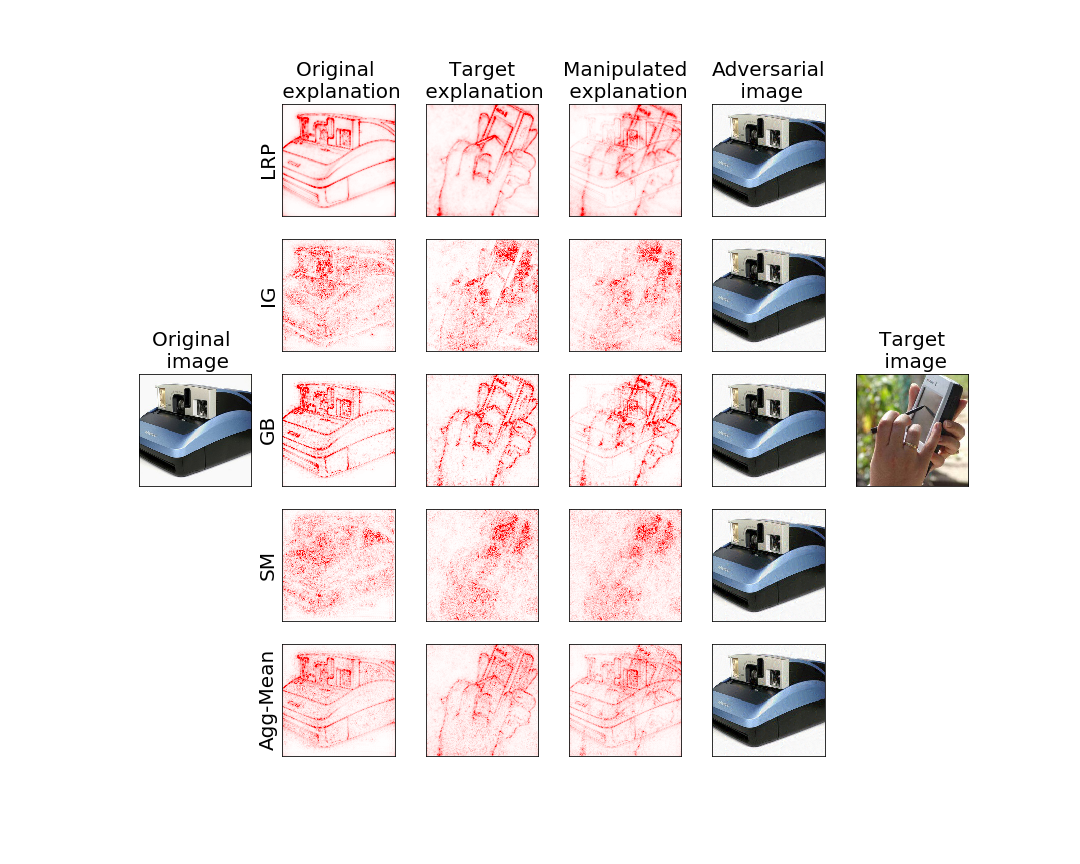}
	\end{center}
	\caption{Appendix example 2. Aggregation is more robust against attack.  There are no visual differences for any of the adversarial inputs.}
	\label{fig:attack3}
\end{figure*}

\begin{figure*}[hbt]
	\begin{center}
		\includegraphics[width=\linewidth]{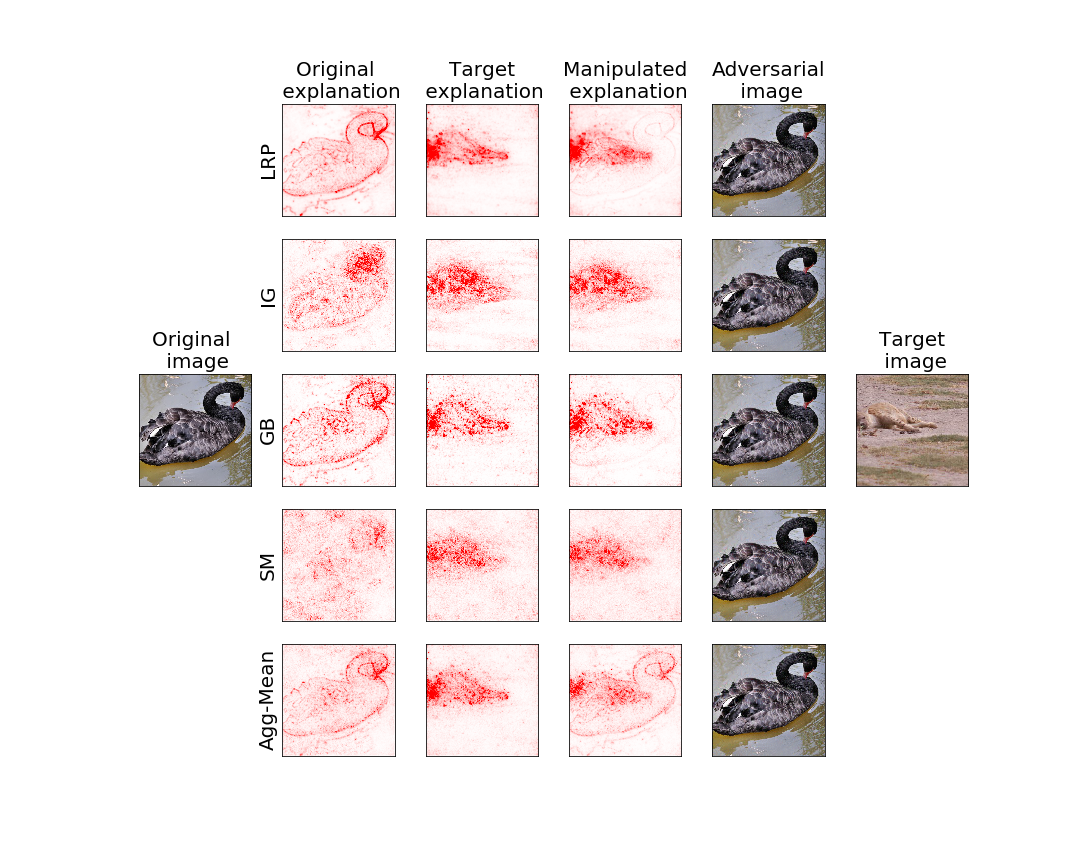}
	\end{center}
	\caption{Appendix example 3. Aggregation is more robust against attack.  There are no visual differences for any of the adversarial inputs.}
	\label{fig:attack2}
\end{figure*}

We provide more examples showing different explanation methods being attacked in \cref{fig:attack0,fig:attack1,fig:attack2,fig:attack3}. An abridged version of \cref{fig:attack0} is also shown in the main text.

\clearpage

\subsubsection{Transferability of attacks}
\begin{figure*}[hbt]
	\begin{center}
		\includegraphics[width=.5\linewidth]{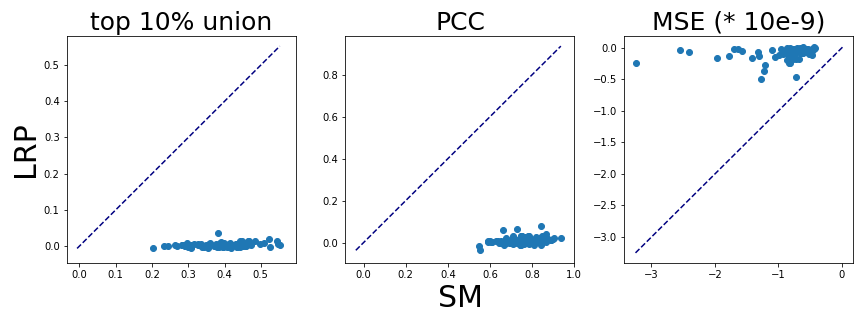}
		\includegraphics[width=.5\linewidth]{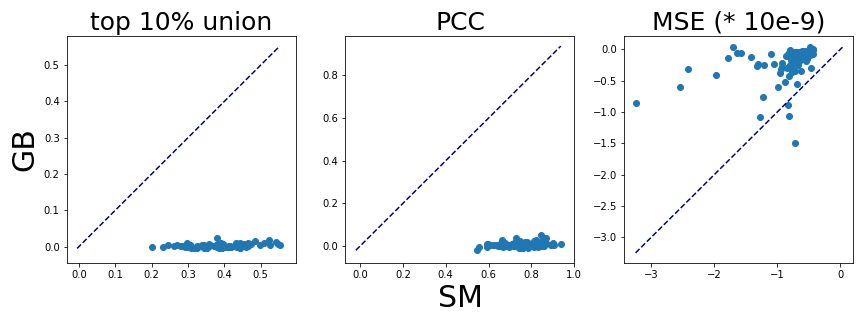}
		\includegraphics[width=.5\linewidth]{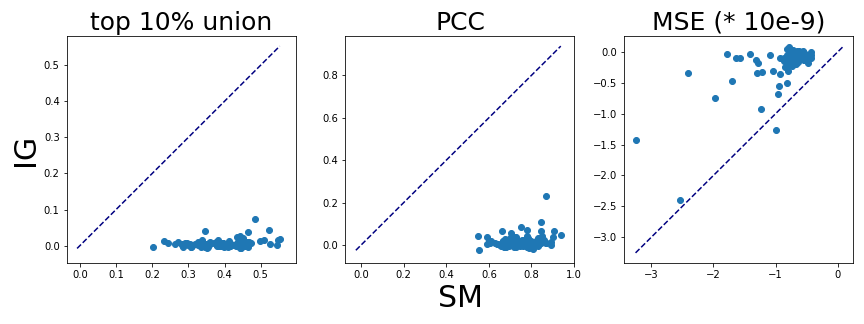}
	\end{center}
	\caption{Gradient as starting method}
	\label{fig:gradient_start}
\end{figure*}

\begin{figure*}[hbt]
	\begin{center}
		\includegraphics[width=.5\linewidth]{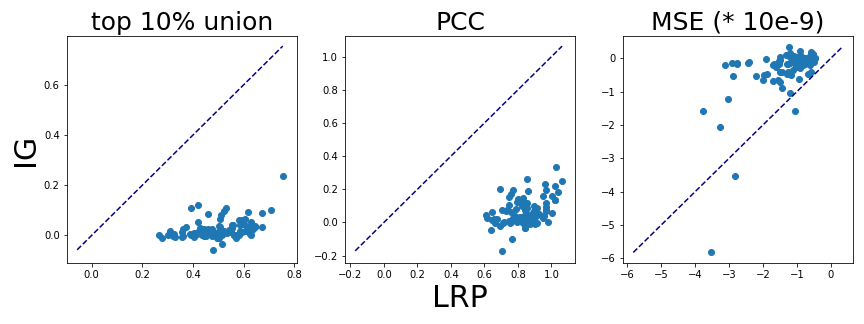}
		\includegraphics[width=.5\linewidth]{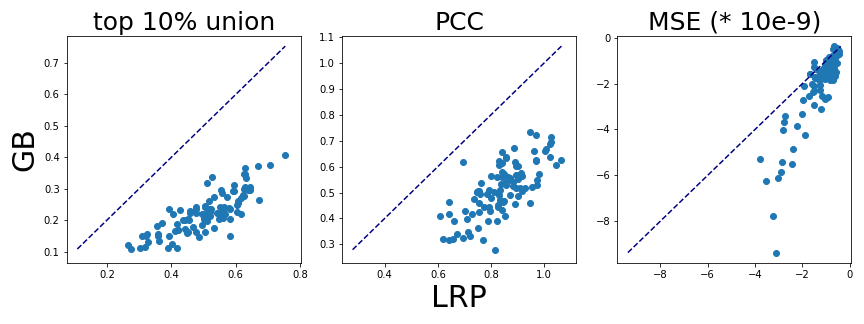}
		\includegraphics[width=.5\linewidth]{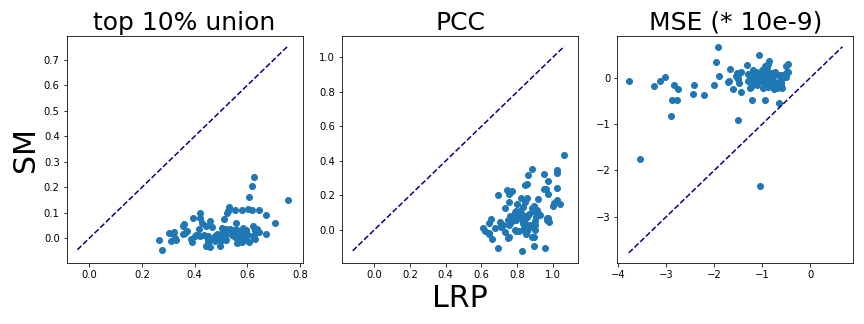}
	\end{center}
	\caption{LRP as starting method}
	\label{fig:lrp_start}
\end{figure*}

\begin{figure*}[hbt]
	\begin{center}
		\includegraphics[width=.5\linewidth]{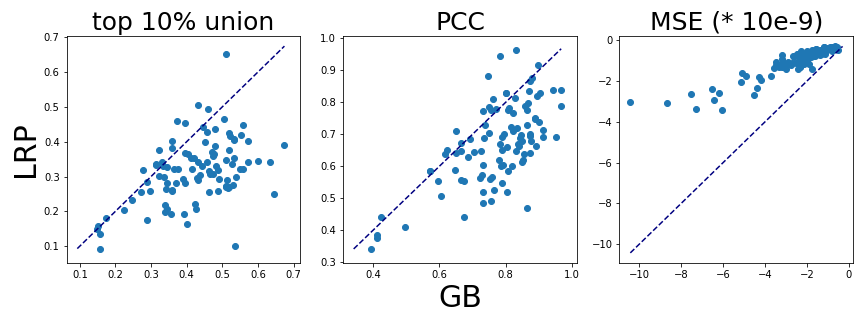}
		\includegraphics[width=.5\linewidth]{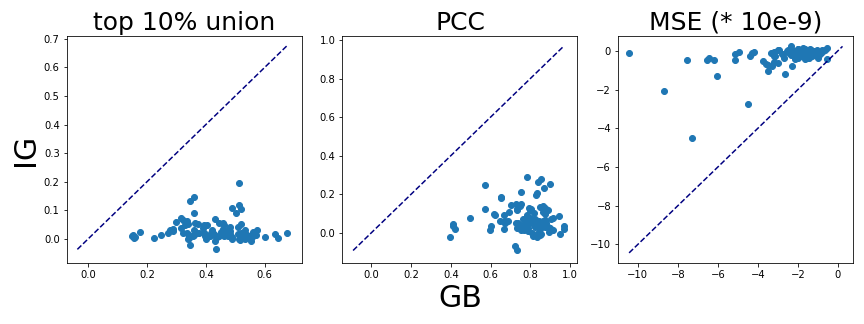}
		\includegraphics[width=.5\linewidth]{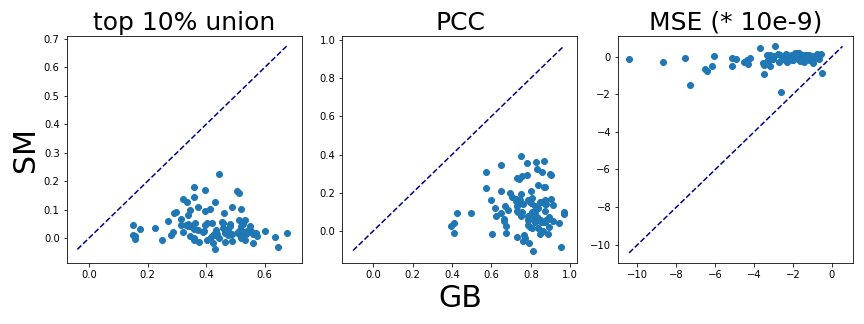}
	\end{center}
	\caption{GuidedBackprop as starting method}
	\label{fig:gb_start}
\end{figure*}

In the main text we show similarity metrics differences between the method being attacked and not being attacked for Guided Backprop and LRP. Here we provide scatter plots for the rest of the considered methods in \cref{fig:ig_start,fig:gradient_start,fig:lrp_start,fig:gb_start}:
\begin{figure*}[hbt]
	\begin{center}
		\includegraphics[width=.5\linewidth]{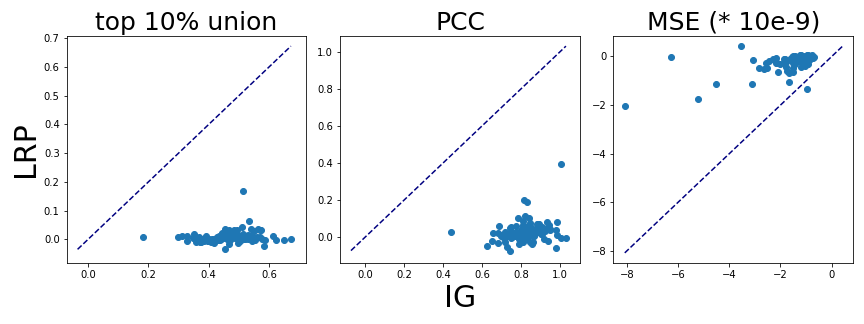}
		\includegraphics[width=.5\linewidth]{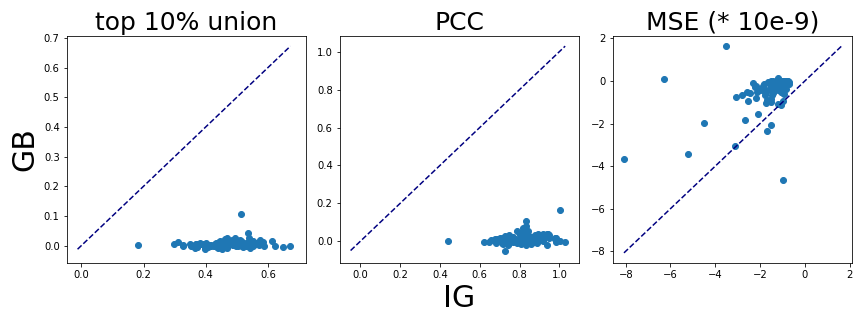}
		\includegraphics[width=.5\linewidth]{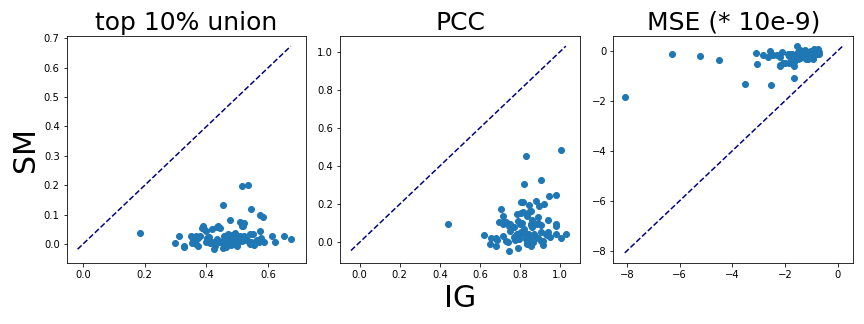}
	\end{center}
	\caption{Integrated Gradient as starting method}
	\label{fig:ig_start}
\end{figure*}

\clearpage
\subsubsection{Similarity of the attacked images to the starting images}
We provide the average distance of the adversarial images to the original images in \cref{tab:results_attacks_extended} (calculated in RGB space, average over all pixels). As can be seen  in \cref{fig:attack0,fig:attack1,fig:attack2,fig:attack3,}, there is no visual difference to the input images for any of the attacks. Attacking \textit{AGG-Mean} has the smallest distance to the input image, supporting our hypothesis that aggregating explanation methods removes vulnerabilities to adversarial manipulation.
\begin{table*}[!hbt]
	\caption{Evaluation scores across methods and architectures on a hundred samples, including similarity of the resulting image to the starting image. Deviation is SE. }
	\label{tab:results_attacks_extended}
	\vskip 0.15in
	\begin{center}
		\begin{small}
			\begin{sc}

				\begin{tabular}{lrrrr}
					\toprule
					{} & MSE $ \Delta $ (*10e-9) & PCC & top 10\% Union &MSE on images \\
					method & & & \\
					\midrule
					SM 		& -0.92 	$\pm$ 0.00				& 0.74 $\pm$ 0.01 			& 0.40 $\pm$ 0.01 & 0.0027 $\pm$ 0.0002\\
					GB 				& -3.25	 $\pm$ 0.02			& 0.77 $\pm$ 0.01			& 0.42 $\pm$ 0.01 &0.0110 $\pm$ 0.0025\\
					LRP & -1.45 $\pm$ 0.01						& 0.81 $\pm$ 0.01 			& 0.49 $\pm$ 0.01  &0.0047 $\pm$ 0.0006\\
					IG 				& -1.76 $\pm$ 0.01			& 0.82 $\pm$ 0.01			& 0.47 $\pm$ 0.01 & 0.0102 $\pm$ 0.0022\\
					AGG-Mean 	& \textbf{-0.89} $\pm$ 0.01 	& \textbf{ 0.54 $\pm$ 0.01 }	& \textbf{ 0.24 $\pm$ 0.01} & 0.0013 $\pm$ 0.0001 \\
					\bottomrule
				\end{tabular}
				
			\end{sc}
		\end{small}
	\end{center}
	\vskip -0.15in
\end{table*}
\clearpage
\subsubsection{Other attacks}
In the main text we mainly concern ourselves with reproducing a pre-specified target explanation, as this is a use case where the motivation of an attacker is apparent. However, as introduced in \citep{ghorbani2019interpretation} other attack objectives are also conceivable. 

\begin{figure*}[hbt]
	\begin{center}
		\includegraphics[width=.7\linewidth]{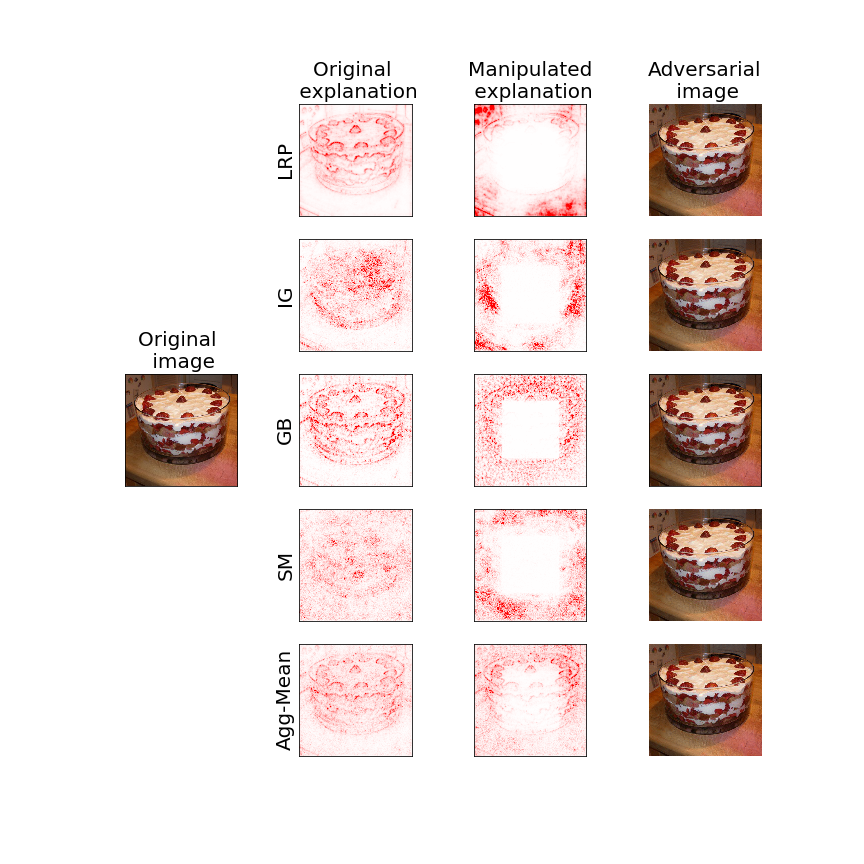}
	\end{center}
	\caption{Attacking explanation methods to make an area irrelevant. AGG-Mean is most robust.}
	\label{fig:square_attack0}
\end{figure*}

\begin{figure*}[hbt]
	\begin{center}
		\includegraphics[width=.7\linewidth]{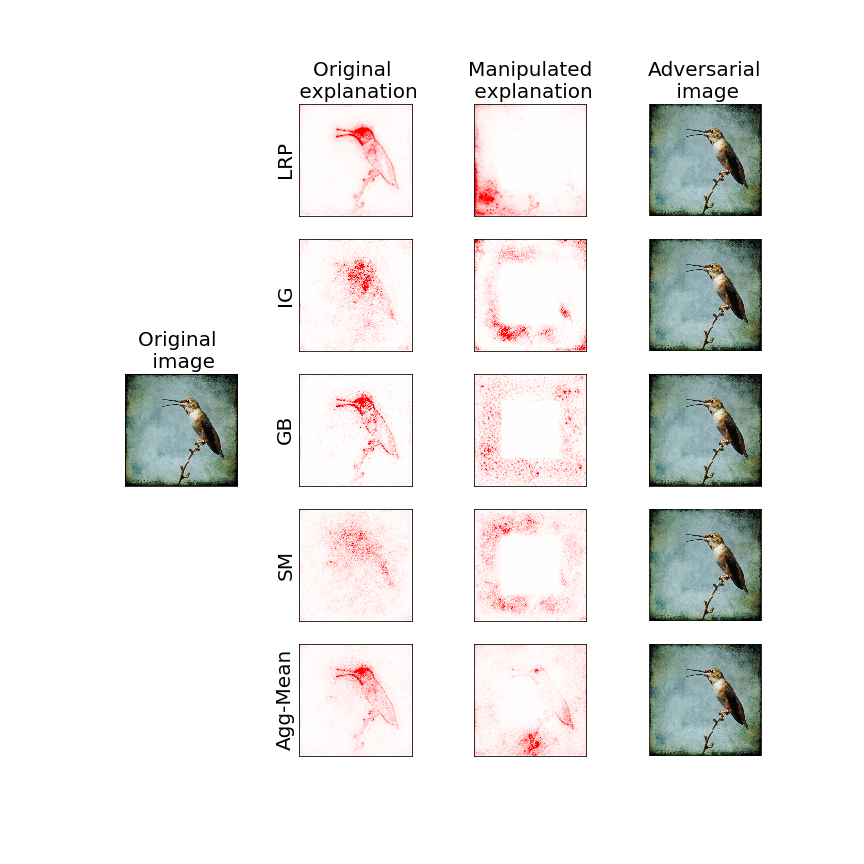}
	\end{center}
	\caption{Attacking explanation methods to make an area irrelevant. AGG-Mean is most robust.}
	\label{fig:square_attack1}
\end{figure*}

We show results when following the objective of making a specified area of the explanation not relevant, i.e.\ a blank space in the explanation as introduced in \citep{ghorbani2019interpretation}. A square (in size a quarter of the image) centered on the middle should not contain any relevance for the explanation. Size and position of the blank space were chosen ad-hoc, we assume that the center of the image generally contains useful information for the classification. We show quantitative results in \cref{tab:square_res}, computing how much percentage of the original relevance is preserved and qualitative results in \cref{fig:square_attack0,fig:square_attack1}. Since the goal of the attack is to entirely remove the relevance in the area, measuring how much relevance is left effectively measures how robust the explanation method is against the adversarial attack.

While an aggregation is not completely robust to the attack, far more of the original explanation is preserved, supporting the results in the main text. 

\begin{table*}[!hbt]
	\caption{Manipulating explanations to show a blank (irrelevant) square. Aggregating explanation methods preserves far more of the original explanation than any single method.  }
	\label{tab:square_res}
	\vskip 0.15in
	\begin{center}
		\begin{small}
			\begin{sc}

				\begin{tabular}{lrrr}
					\toprule
					{} &  start\_percentage &  end\_percentage &  preserved \\
					method          &                   &                 &            \\
					\midrule
					SM        &              0.09 &        1.84e-02 &       0.20 \\
					GB &              0.11 &        2.58e-03 &       0.02 \\
					IG &              0.11 &        2.31e-02 &       0.21 \\
					LRP             &              0.11 &        1.71e-03 &       0.02 \\
					agg-mean     &              0.10 &       \textbf{ 3.09e-02} &       \textbf{0.31} \\
					\bottomrule
				\end{tabular}

			\end{sc}
		\end{small}
	\end{center}
	\vskip -0.15in
\end{table*}


\end{document}

%% file: math_commands.tex
\usepackage{amsmath,amsfonts,bm}

\def\eqref#1{equation~\ref{#1}}

\def\1{\bm{1}}

\DeclareMathAlphabet{\mathsfit}{\encodingdefault}{\sfdefault}{m}{sl}
\SetMathAlphabet{\mathsfit}{bold}{\encodingdefault}{\sfdefault}{bx}{n}

\newcommand{\R}{\mathbb{R}}



%% file: tables/AUC.tex
\begin{tabular}{lrrrrr}
	\toprule
	Method              &     Inception &  ResNet101 &  ResNet50 &  VGG19 &  Xception \\
	\midrule
	LIME                     &       74.0 &       67.6 &      66.7 &   73.4 &      74.0 \\
	SM                       &       76.2 &       75.6 &      76.2 &   81.4 &      72.6 \\
	GB                       &       73.7 &       77.6 &      80.9 &   84.3 &      74.8 \\
	IG                       &       76.0 &       76.0 &      76.2 &   79.8 &      73.3 \\
	SG                       &       77.9 &       77.7 &      77.5 &   83.9 &      75.0 \\
	GC                       &       78.7 &       77.8 &      78.5 &   86.1 &      73.5 \\
	Agg-Mean 				 &       79.9 &       81.1 &      80.9 &   86.6 &      \textbf{76.7 }\\
	Agg-Var   				 &       \textbf{80.4 }&      \textbf{ 81.2 }&      \textbf{81.0} &   \textbf{86.7} &     \textbf{ 76.7 }\\
	\bottomrule
\end{tabular}

%% file: tables/jaccard.tex
\begin{tabular}{lrrr}
\toprule
	{} &  ResNet101 &  ResNet50 &  VGG19 \\
	Method      &            &           &        \\
	\midrule
	AGG-Mean    &       0.63 &      0.66 &   0.64 \\
	AGG-Var     &       0.66 &      0.68 &   0.67 \\
	GB          &       0.42 &      0.49 &   0.47 \\
	GC          &       0.60 &      0.62 &   0.60 \\
	IG          &       0.45 &      0.45 &   0.47 \\
	Mean(SG+GC) &       \textbf{0.69} &      \textbf{0.70} &   \textbf{0.65} \\
	SG          &       0.63 &      0.65 &   0.59 \\
	SM          &       0.45 &      0.45 &   0.47 \\
\bottomrule
\end{tabular}